\documentclass[review]{elsarticle}

\usepackage{etoolbox}
\makeatletter
\patchcmd{\ps@pprintTitle}
  {Preprint submitted to}
  {}
  {}{}
\makeatother

\usepackage[utf8]{inputenc}
\usepackage{amsfonts}
\usepackage{color}
\usepackage{xcolor}

\usepackage{graphicx}
\graphicspath{ {resources/} }

\usepackage{algorithm}
\usepackage{algpseudocode}

\usepackage{amsmath}

\usepackage{adjustbox}
\usepackage{rotating}

\journal{Technical report}

\begin{document}

\begin{frontmatter}

\title{A Hybrid Genetic Algorithm for the Traveling Salesman Problem with Drone}

\author{Quang Minh Ha}
\address{ICTEAM, Université Catholique de Louvain, Belgium}
\address{quang.ha@uclouvain.be}

\author{Yves Deville}
\address{ICTEAM, Université Catholique de Louvain, Belgium}
\address{yves.deville@uclouvain.be}

\author{Quang Dung Pham}
\address{SOICT, Hanoi University of Technology, Vietnam}
\address{}

\author{Minh Hoàng Hà}
\address{minhhoang.ha@vnu.edu.vn}
\address{ORLab, University of Engineering and Technology, Vietnam National University, Vietnam}

\begin{abstract}
This paper addresses the Traveling Salesman Problem with Drone (TSP-D), in which a truck and drone are used to deliver parcels to customers. The objective of this problem is to either minimize the total operational cost (min-cost TSP-D) or minimize the completion time for the truck and drone (min-time TSP-D). This problem has gained a lot of attention in the last few years since it is matched with the recent trends in a new delivery method among logistics companies. To solve the TSP-D, we propose a hybrid genetic search with dynamic population management and adaptive diversity control based on a split algorithm, problem-tailored crossover and local search operators, a new restore method to advance the convergence and an adaptive penalization mechanism to dynamically balance the search between feasible/infeasible solutions. The computational results show that the proposed algorithm outperforms existing methods in terms of solution quality and improves best known solutions found in the literature. Moreover, various analyses on the impacts of crossover choice and heuristic components have been conducted to analysis further their sensitivity to the performance of our method.
\end{abstract}


\end{frontmatter}

\section{Introduction}

The past few years have witnessed a rapid growth of interest in research on a novel approach for delivering parcels to customers, which is traditionally handled by land vehicles such as trucks. This new method utilizes drones with trucks to not only reduce delivery time and operational cost but also improve service quality. A problem related to this new delivery method is called the routing problem with drones, which is a generalization of the well-known Traveling Salesman Problem (in the case of one truck and one drone) and Vehicle Routing Problem (in the case of a fleet of trucks and drones); they are denoted TSP-D and VRP-D, respectively, and their objective is to minimize either the total operational cost (min-cost) or the completion time for a truck and drone (min-time).

In the literature, the very first work on this class of problems is the work of Murray and Chu \cite{murray2015flying}, in which the authors proposed two subproblems. The first problem is a TSP-D problem in which a truck and drone cooperate with each other to deliver parcels. The authors named it the ``Flying Sidekick Traveling Salesman Problem'' (FSTSP) and introduced a mixed integer programming formulation and a simple and fast heuristic with the objective of minimizing the completion time for two vehicles. The second problem is called the ``Parallel Drone Scheduling TSP'' (PDSTSP), in which a single truck and a fleet of drones are in charge of delivering parcels. The truck is responsible for parcels far from the distribution centre (DC), and the drones are responsible for serving customers in its flight range around the DC. Again, the objective is to minimize the latest time that a vehicle returns to the depot. The problem description and hypothesis used in FSTSP has been adapted in numerous subsequent studies such as in \cite{ha2018min}, \cite{ponza2016optimization}, and \cite{freitas2018variable} as well as in this paper.

Agatz et al. \cite{agatz2018optimization} also introduced a TSP-D problem with assumptions differing from those of the FSTSP. The most notable is that the drone may be launched and returned to the same location (whereas this is forbidden in FSTSP). Additionally, the two vehicles share the same road network (they are in different networks in FSTSP). The authors proposed a mathematical model for this problem and developed several route-first, cluster-second heuristics based on local search and dynamic programming to solve it with instances with up to 10 customers. The above work has been extended further by Bouman et al. \cite{bouman2018dynamic}, who presented exact solution approaches, proving that the problem with larger instances can be solved.

In a recent work, Freitas et al. \cite{freitas2018variable} proposed a hybrid heuristic named HGVNS to solve two TSP-D variants by \cite{murray2015flying} and \cite{agatz2018optimization} with the min-time objective. In detail, HGVNS first obtains the initial solution by using an MIP solver to solve the TSP optimally and then applies a heuristic in which some trucks' customers are removed and reinserted as drone customers. Next, the initial solution is used as the input for a general variable neighbourhood search in which eight neighbourhoods are shuffled and chosen randomly. The authors conducted the experiments on three instance sets from \cite{ponza2016optimization}, \cite{agatz2018optimization} and TSPLIB. The computational results show that the proposed approach can decrease delivery time by up to 67.79\%.

A generalization of the TSP-D is firstly studied by Wang et al. \cite{wang2017vehicle} where a fleet of trucks and drones is responsible for delivering parcels. The authors named it the ``Vehicle Routing Problem with Drones'' (VRPD or VRP-D). Several theoretical aspects have been studied in terms of bounds and worst cases. An extension of that work was studied in \cite{poikonen2017vehicle}, in which the author considered more practical aspects such as drone endurance and cost. In addition, connections between VRPD and other classes of VRPs have been made in the form of bounds and asymptotic results.

Other works related to drone applications are also presented in a survey conducted by Otto et al. \cite{otto2018optimization}.

In this paper, we introduce a new \textit{hybrid genetic algorithm (HGA) with adaptive diversity control} to effectively solve the TSP-D under both min-cost and min-time objectives. HGA is a combination of the genetic algorithm and local search technique together with a population management, diversity control and penalization mechanism to balance the search between feasible and infeasible search spaces. This method was initially proposed by \cite{vidal2012hybrid} and has been used to solve many variants of VRP efficiently, as in \cite{vidal2012hybrid}, \cite{vidal2013hybrid}, \cite{vidal2014unified}, and \cite{Bulhoes}. We also present problem-tailored components to significantly facilitate the performance of the algorithm. Different computational experiments show the improvements in terms of solution quality under both objectives and different instance sets and the importance of the new proposed elements. 

The main contributions of this paper are as follows.
\begin{itemize}
	\item We propose an efficient hybrid genetic algorithm that includes a new crossover, a set of 16 local search operators, and a penalization and restore mechanism to solve the TSP-D under both min-cost and min-time objectives.
	\item We conduct extensive computational experiments to evaluate the performance of HGA under instance sets from \cite{murray2015flying} and \cite{ha2018min}. The proposed method outperforms existing approaches in terms of solution quality and can improve a number of best known solutions.
	\item We analyze the efficiency and importance of the new components to the performance of the overall algorithm.
\end{itemize}

The remaining parts of the paper are organized as follows. Section \ref{section:problem-description} introduces the TSP-D and related assumptions considered in the problem. Section \ref{section:memetic} discusses the proposed hybrid genetic algorithm (HGA). Section \ref{section:computational-results} presents the computational results, and Section \ref{section:conclusion} concludes the paper.

\section{Problem description}
\label{section:problem-description}
In this section, we briefly discuss the description of the TSP-D, which was first proposed in \cite{murray2015flying} and then developed further in \cite{ha2018min} to solve the min-cost objective. In this problem, given a graph $G = (V, A), V = \{0 \ldots n + 1\}$ is a set of depot and customer locations and $A$ is a set of arcs that link two pair of nodes in $V$. We need to deliver parcels to a set $N = \{1, \ldots n\}$ customers using a truck and a drone (an unmanned aerial vehicle used for delivery). In this graph, $0$ is the depot, and $n + 1$ is its duplication. We denote $d_{ij}$ and $\tau_{ij}$ ($d'_{ij}$ and $\tau'_{ij}$) as the distance and time traveled from node $i$ to node $j$ by truck (drone), respectively. The effective arrival times of the truck and drone are denoted by $t_i$ and $t'_i$. We have $t_0 = t'_0 = 0$. Different from actual arrival time, the effective arrival time of a vehicle (drone or truck) takes into account both the actual arrival time and the time required to retrieve and (possibly) prepare the drone for the next launch. This definition was initially described in the work of \cite{murray2015flying}.
The drone is managed by the truck driver and is carried in the truck while not in service. To make a delivery, the drone is launched from either the truck or the starting depot and later returns to the truck or the return depot. The launch and return locations must be different locations. The delivery plan of these two vehicles (truck and drone) is subjected to the following requirements.
\begin{itemize}
	\item Both vehicles (truck and drone) must start from and return to the depot.
	\item Each customer can only be serviced once by either a truck or drone. If a customer is served by a truck (a drone), we call it a \textit{truck delivery} (\textit{drone delivery}).
	\item A drone delivery is represented as a 3-tuple $\langle i, j, k \rangle$, where $i, j, k$ are customer locations that are described as follows.
	\begin{itemize}
		\item $i$ is the node where the truck launches the drone, which we call the \textit{launch node}. We also denote $s_L$ as the time required for the truck driver to prepare the drone for launch.
		\item $j$ is the node the drone will fly to and make the delivery. We call it the \textit{drone node}. Most importantly, node $j$ must be eligible for the drone to visit, as not all parcels can be delivered by the drone due to capacity limitation (i.e., the parcel is too heavy). We denote the set of nodes that can be served by drone as $V_D \subseteq N$.
		\item $k$ is the rendezvous node, where the drone -- after making a delivery -- rejoins the truck to have its battery recharged and to be made ready for the next launches. The time required for those actions is denoted $s_R$. In addition, two vehicles are required to \textbf{wait for each other} at the rendezvous point, and while waiting for the truck, the drone is assumed to be in constant flight.
	\end{itemize}
	\item In a drone delivery, both truck and drone are required to satisfy the endurance constraint, which is, in detail:
	\begin{itemize}
		\item \textbf{Truck travel time constraint:} the truck travel time from the launch node to the rendezvous node plus its recovery time cannot exceed the \textbf{drone endurance} (the maximum operational time of a drone without recharging), 
			\begin{equation}
				\tau_{i \rightarrow k} + recover_{truck} \leq \epsilon \nonumber
			\end{equation}
where $\tau_{i \rightarrow k}$ is the truck travel time from $i$ to $k$, and $recover_{truck}$ is the time taken for the truck to recover the drone and possibly prepare it for the next launch. More specifically, if the truck just recovers the drone without relaunching it at the same location, then $recover_{truck} = s_R$. Otherwise, if the truck relaunches the drone at the same location, then $recover_{truck} = s_R + s_L$. \textbf{This constraint is not imposed when the launch node is the depot (node 0)}.
		\item \textbf{Drone travel time constraint:} the drone travel time plus its recovery time cannot exceed the drone endurance:
			\begin{equation}
				\tau'_{ij} + \tau'_{jk} + recover_{drone} \leq \epsilon \nonumber
			\end{equation}
where $recover_{drone} = s_R$ is the time taken to recover the drone.
	\end{itemize}	 
	\item We denote $\mathcal{P}$, the set of all possible drone deliveries, as follows:
	\begin{equation}
		\mathcal{P} = \{ \langle i, j, k \rangle : i, k \in V, j \in V_D, i \neq j \neq k, \tau'_{ij} + \tau'_{jk} \leq \epsilon \}, 
	\end{equation}
where $\epsilon$ is the drone endurance.
	\item Each vehicle has its own transportation cost per unit of distance, denoted $\mathcal{C}_1$ and $\mathcal{C}_2$ for the truck and drone, respectively.
	\item When two vehicles have to wait for each other at the rendezvous point, waiting costs are created and added to the transportation cost to form the total operational cost of the system. These waiting costs are calculated as
	\begin{equation}
		w_T = \alpha \times wt_T, \text{and}
	\end{equation}
	\begin{equation}
		w_D = \beta	\times wt
		_D, 
	\end{equation}
where $wt_T, wt_D$ are the waiting times; $w_T, w_D$ are the waiting costs of the truck and drone, respectively, and $\alpha, \beta$ are the waiting fees of two vehicles per unit time.
\end{itemize}

The objectives of the TSP-D are either to minimize the total operational cost of the system or to minimize the completion time of two vehicles. We denote the problem with the first objective as ``min-cost TSP-D'' and with the latter as ``min-time TSP-D''.

\section{A Hybrid Genetic Algorithm for TSP-D (HGA)}
\label{section:memetic}

In this section, we describe a hybrid genetic algorithm with adaptive diversity control method for solving TSP-D. The framework, as proposed in \cite{vidal2012hybrid}, is a hybrid metaheuristic that combines the exploration capability of genetic algorithms with efficient local search and diversity control. We adapt this general framework with modifications to match the characteristics of the TSP-D. They include new local search operators, crossovers, a penalized mechanism and a restoration method to convert from a TSP-D solution to a giant-tour chromosome. We describe the approach in Algorithm \ref{alg:memetic}. In detail, starting from an initial population (Line 1), for each iteration, two parents are selected to generate an offspring individual using a crossover operator (Line 4). This offspring then goes through a split procedure (proposed in \cite{ha2018min}) to obtain the drone delivery and truck delivery chromosome (Line 5). Subsequently, the offspring is ``educated'' by a local search method -- which contains multiple operators -- to improve its quality. The educated offspring then employs a restore method to update its giant tour chromosome (Line 7). The offspring is then checked for feasibility and is added to the appropriate subpopulation (feasible or infeasible). It also has a probability of being repaired of 50\% and is added to a feasible subpopulation if the repair succeeds (Lines 8 to 14). In the next step, if a subpopulation reaches its maximum size, a survivor selection method is called to eliminate a number of individuals in that subpopulation, keeping only the best ones (Lines 15 to 17). The method then adjusts the penalty parameters (Line 18) and calls the \textit{diversification} procedure if the search is not improved after a certain number of iterations (Line 19). Finally, we return the best feasible solution found (Line 22).

\begin{algorithm}[H]
\caption{HGA for TSP-D}\label{alg:memetic}
\begin{algorithmic}[1]
\State Initialize population
\While{number of iterations without improvement $<$ $Iter_{NI}$}
	\State Select parents $P1$ and $P2$
	\State Generate offspring individual $C$ from $P1$ and $P2$
	\State Apply split on $C$
	\State Educate $C$ using local search
	\State Call restore method to update the giant-tour chromosome in $C$
	\If {$C$ is infeasible}
		\State insert $C$ into infeasible subpopulation
		\State with probability $P_{rep}$, repair $C$
	\EndIf	
	\If {$C$ is feasible}
		\State insert $C$ into feasible subpopulation
	\EndIf
	\If {maximum subpopulation size reached}
		\State Select survivors
	\EndIf
	\State Adjust the penalty parameters for violating the drone endurance constraint
	\If {best solution is not improved for $Iter_{DIV}$ iterations}
		Diversify population
	\EndIf
\EndWhile
\State Return the best feasible solution
\end{algorithmic}
\end{algorithm}

The rest of this section is arranged as follows. We first define the search space in Section \ref{section:memetic-search-space}. Section \ref{section:memetic-solution-representation} describes the solution representation. Section \ref{section:memetic-evaluation} presents the evaluation of individuals. Parent selection and crossover are described in Section \ref{section:memetic-crossover}. Section \ref{section:memetic-localsearch} discusses the local search procedure, and various operators are presented. The restore method is introduced in Section \ref{section:memetic-restore}. Finally, Section \ref{section:population-management} regards the population management with the population initialization, adjustment of penalty coefficients, survivor selection and diversity control.

\subsection{Search space}
\label{section:memetic-search-space}

It has been well studied that by exploiting infeasible solutions, we can significantly improve the performance of a heuristic \cite{glover2011case}. In this section, we define the search space $\mathcal{S}$, which includes the feasible and infeasible solutions $s \in \mathcal{S}$. Infeasible solutions comprise drone deliveries that violate the drone endurance constraint. More specifically, a drone delivery in a TSP-D solution is not valid in the following two scenarios.
\begin{itemize}
	\item The truck travel time constraint is violated (except for the case where the launch node is the depot, as described in the problem description above):
	\begin{equation}
		\nonumber
		\tau_{i \rightarrow k} + recover_{truck} > \epsilon, 
	\end{equation}
where $\tau_{i \rightarrow k}$ is the truck travel time from $i$ to $k$.
	\item The drone travel time constraint is violated:
	\begin{equation}
		\nonumber
		\tau'_{ij} + \tau'_{jk} + recover_{drone} > \epsilon.
	\end{equation}
\end{itemize}

In these two cases, the drone cannot feasibly be flown because its battery will be depleted before the retrieval operation undertaken by the truck driver is completed.

Let $sol(s)$ represent the TSP-D solution in the search space. We have $sol(s) = (TD, DD)$, where $TD = \langle e_0, \ldots e_k \rangle$ is the truck tour, and $DD \subseteq \mathcal{P}$ is the set of drone deliveries in solution $s$.

We now define the fitness evaluation function for min-time and min-cost TSP-D separately.

\subsubsection{Min-cost TSP-D}
For min-cost TSP-D, the operational cost of solution $s$, denoted $cost(TD, DD)$, is calculated as
\begin{equation}
	cost(TD, DD) = cost(TD) + cost(DD) + cost_W(DD),
\end{equation}
where
\begin{itemize}
	\item[-] $cost(TD) = \sum\limits_{e = 0}^{k - 1}\mathcal{C}_1 d_{i, i + 1}$ is the cost of the truck tour;
	\item[-] $cost(DD) = \sum\limits_{\langle i, j, k \rangle \in DD}\mathcal{C}_2 (d'_{ij} + d'_{jk})$ is the cost of drone deliveries; and
	\item[-] $cost_W(DD) = \sum\limits_{\langle i, j, k \rangle \in DD}cost_W^T(\langle i, j, k \rangle) + cost_W^D(\langle i, j, k \rangle$ is the wait cost of the truck and drone. We have $cost_W^T = \alpha \times max(0, \tau_{i \rightarrow k} - \tau'_{ijk})$, where $\tau_{i \rightarrow k}$ is the truck travel time from $i$ to $k$ (in the truck tour), and $\tau'_{ijk}$ is the drone travel time from $i$ to $j$ to $k$. In addition, $cost_W^D = \beta \times max(0, \tau'_{ijk} - \tau_{i \rightarrow k})$.
\end{itemize}

Let $\omega_C$ represent the penalty for violating the drone endurance constraint. We define the \textbf{penalized cost} of a solution $s$ as the sum of the operational cost and the weighted sum of the truck's or drone's excess travel time during drone deliveries. This \textbf{penalized cost} is computed as
\begin{equation}
\begin{split}
	\phi(s) = cost(TD, DD) + \omega \sum\limits_{\langle i, j, k \rangle \in DD} & max(0, \tau_{i \rightarrow k} + recover_{truck} - \epsilon) \times \Upsilon_T \times \mathcal{C}_1 + \\
	& max(0, \tau'_{ij} + \tau'_{jk} + recover_{drone}  - \epsilon) \times \Upsilon_D \times \mathcal{C}_2,
\end{split}
\end{equation}

where $\omega$ is the penalty for violating the constraint, and $\Upsilon_T$ and $\Upsilon_T$ are the speeds of the truck and drone, respectively. This penalized cost is then used as the fitness function to compute the fitness of the individuals.

\subsubsection{Min-time TSP-D}
In min-time TSP-D, the completion time of a solution $s$, denoted $time(s)$, is calculated as
\begin{equation}
	time(s) = max(t_{n+1}, t'_{n+1}).
\end{equation}
Similar to min-cost TSP-D, we also have the penalized cost of a solution $s$ in the min-time objective as the sum of the completion time of two vehicles and the penalties for violating the constraint. It is computed as follows:
\begin{equation}
\begin{split}
	\phi(s) = time(s) + \omega \sum\limits_{\langle i, j, k \rangle \in DD} max(0, max(\tau_{i \rightarrow k} + recover_{truck}, \\ 	
	\tau'_{ij} + \tau'_{jk} + recover_{drone}) - \epsilon).
\end{split}
\end{equation}

Again, this is used to compute the fitness of individuals.

\subsection{Solution representation}
\label{section:memetic-solution-representation}

\begin{figure}[H]
\center
\includegraphics[scale=0.5]{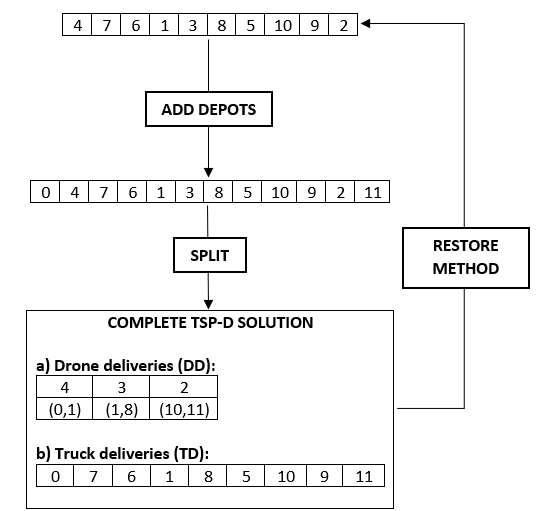}
\caption{Transformations between giant-tour chromosome and TSP-D solution}\label{figure:solution-representation}
\end{figure}

A solution in HGA is represented as a giant TSP tour (giant tour) with two depots removed. We also denote this as a (giant-tour) \textbf{chromosome}. When a TSP-D solution is needed for a local search method, it can be obtained using the \textit{split} procedure, which runs in polynomial time \cite{ha2018min}. Reversely, we can retrieve a giant tour from a TSP-D solution by using a \textit{restore method}, which will be discussed in the coming section. To conclude, by having a transformation between a giant tour and TSP-D solution using the split and restore method, we can use the fast and efficient operators in both the crossover and local search step. A demonstration of this transformation is described in Figure \ref{figure:solution-representation}.

\subsection{Individual evaluation}
\label{section:memetic-evaluation}

To evaluate any individual $P_1$ in the population, we consider two factors: its penalized cost $\phi(P_1)$ (described in Section \ref{section:memetic-search-space}) and its contribution to the diversity of the population, denoted $\Delta(P_1)$ and calculated as the average ``distance'' from $P_1$ to its closest neighbours in the population. By taking into account these two factors, we can obtain a balance between intensification and diversification. Otherwise, the heuristic might either converge too soon and too quickly (focusing only on improving the penalized cost) or will always explore completely different giant tours, leading to a random search. In detail, the diversity contribution described above is presented in Equation \ref{equation: bf}:
\begin{equation} \label{equation: bf}
	\Delta(P_1) = \frac{1}{n_{close}}\sum\limits_{P_2 \in \mathcal{N}_{close}} \delta(P_1, P_2), 
\end{equation}
where $n_{close}$ is the number of considered closest neighbours, and $\mathcal{N}_{close}$ is the set of closest neighbours of $P_1$ (i.e., the set of elements sorted using Equation $\ref{equation:bf-distance}$). The distance between two individuals $P_1$ and $P_2$, denoted $\delta(P_1, P_2)$, is a normalized Hamming distance based on the differences between the nodes in the same positions of the giant-tour chromosome. This distance is shown in Equation \ref{equation:bf-distance}, where $\textbf{1}(condition)$ is a valuation function that returns 1 if the condition is true and 0 otherwise. 

\begin{align} \label{equation:bf-distance}
\delta(P_1, P_2) = \frac{1}{n}\sum\limits_{i = 1, \ldots, n}[\textbf{1}(P_1.gt(i) \neq P_2.gt(i))],
\end{align}
where $\textbf{1}(P_1.gt(i) \neq P_2.gt(i))$ returns 1 if the giant-tour chromosome in $P$ contains a different node to the giant-tour chromosome in $P_2$ in the same position $i$ and 0 otherwise.

The evaluation of an individual $P$, or as we call it, the \textbf{biased fitness}, denoted $BF(P)$, is then computed as in Equation \ref{equation:evaluation}, where $fit(P)$ is the rank of $P$ in the subpopulation of size $nbIndiv$ with respect to its penalized cost $\phi(P)$, and $dc(P)$ is the rank of $P$ in the subpopulation in terms of diversity contribution. The parameter $nbElite$ ensures that a certain number of elite individuals will survive to the next generation during the survival selection process (proven in \cite{vidal2012hybrid}).

\begin{equation} \label{equation:evaluation}
	BF(P) = fit(P) + (1 - \frac{nbElite}{nbIndiv} dc(P) )
\end{equation}

\subsection{Parent selection and crossover}
\label{section:memetic-crossover}

Each iteration in HGA includes a generation of a new child chromosome. This is done by first merging two subpopulations into one population and randomly selecting two parents, $P_1$ and $P_2$, in that population using the tournament selection method. In detail, to choose a parent, we pick two individuals from the complete population above and select the one with the best biased fitness. Two parents have then gone through a crossover step.

\begin{figure}[H]
\center
\includegraphics[scale=0.65]{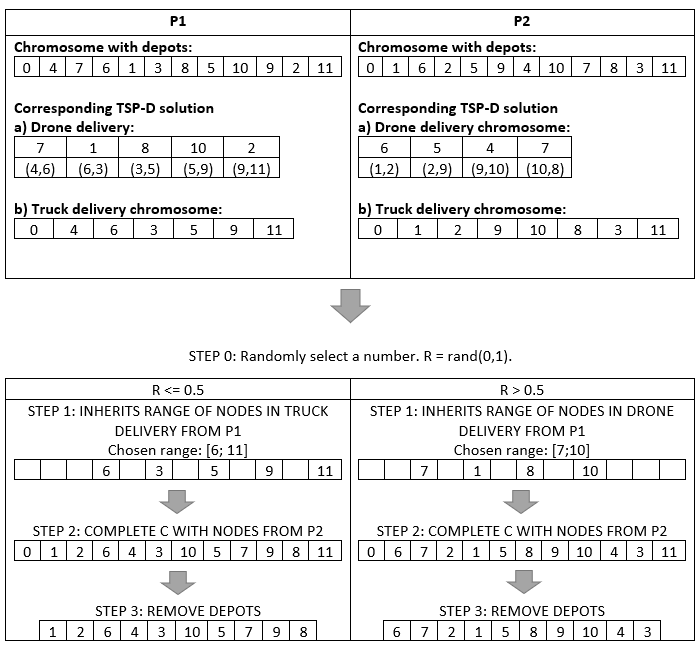}
\caption{DX Crossover for TSP-D.}\label{figure:dx}
\end{figure}

For crossover operators, one can use the classical TSP crossovers -- OX (order crossover), PMX (partially mapped crossover), OBX (order-based crossover), and PBX (position-based crossover) \cite{potvin1996genetic}. In this paper, we propose a problem-dependent crossover called DX that can solve the TSP-D more effectively. The most important feature of DX is that it exploits the characteristics of a TSP-D solution -- the drone deliveries and truck deliveries -- and try to transmit that information from the parents to the offspring. A detailed description of this crossover is presented in Figure \ref{figure:dx}, and the crossover is described in Algorithm \ref{alg:crossover-dx}.

\begin{algorithm}[H]
\caption{Crossover DX for TSP-D}\label{alg:crossover-dx}
\begin{algorithmic}[1]
\State \textbf{Input: Parents $P_1, P_2$ and the corresponding TSP-D solution of $P_1$ which is $(TD_1, DD_1)$}
\State Let $TSP_1$ = $P_1$ with 2 depots added, $TSP_2$ = $P_2$ with 2 depots added
\State Let $C$ = An empty chromosome with 2 depots added
\State Let $r$ = A random number in range of [0, 1];
\If {$r \leq 0.5$}
	\State Choose 2 cut points $a, b, a < b$ in $TD_1$ and copy the nodes between these cut points to $C$ while respecting its position in $TSP_1$
\Else
	\State Choose 2 cut points $a, b, a < b$ in $DD_1$ and copy the nodes between these cut points to $C$ while respecting its position in $TSP_1$
\EndIf
\State Fill the other positions of $C$, starting at position 1, by taking the remaining nodes of $TSP_1$ while keeping their relative orders in $TSP_2$.
\State Return $C$ with 2 depots removed.
\end{algorithmic}
\end{algorithm}

In detail, Algorithm \ref{alg:crossover-dx} first takes the two parents $P_1, P_2$ as one of its inputs. Moreover, in line 1, it also takes into account the corresponding TSP-D solution $(TD_1, DD_1)$ of $P_1$, which was obtained during the ``education'' process (Line 6 of Algorithm \ref{alg:memetic}). Subsequently, it defines two TSP tours, $TSP_1, TSP_2$, in Line 2 by taking two parents and adding two depots to them. An empty offspring with two depots is also initialized in Line 3. In Line 4, a random number is generated to decide from which component -- $TD_1$ or $DD_1$ -- the algorithm will inherit. In either case, it will choose a random segment of the chosen component by generating two random cut points, $a, b$, with $a < b$, and copy the nodes between those cut points to C while keeping their original positions in $TSP_1$ (Lines 5 to 9). Finally, the remaining nodes of $C$ are filled one by one, starting from position 1, by taking the remaining nodes of $TSP_1$ and copying to $C$ while keeping their relative orders in $TSP_2$ (Line 10). The offspring is returned by removing two depots of $C$ (Line 11).

\subsection{Education using local search}
\label{section:memetic-localsearch}

The main role of the \textit{education} step is to improve the quality of solutions by means of the local search procedure. We design a hill-climbing and first-improvement local search for both min-cost and min-time objectives. Similar to \cite{vidal2012hybrid}, we also apply the technique proposed in \cite{toth2003granular} to restrict the search to $h \times n$ closest vertices, where $h = 0.1$ is the \textit{granular threshold}. This technique allows to reduce significantly the computation time consumed by the education process. A set of 16 move operators is proposed to explore the neighbourhoods of TSP-D. In each operator, the evaluation separately evaluates the move costs for the min-cost and min-time objectives. For min-cost, it is the total truck and drone costs of the affected arcs, while the total truck and drone travel times of the affected arcs are calculated in the min-time problem. Moreover, the truck and drone cumulative time and cost as well as the cost and time of all drone tuples in set $\mathcal{P}$ are pre-computed at the beginning of the HGA to effectively accelerate the algorithm.

We now describe in detail the neighbourhoods to be explored.
\begin{itemize}
	\item[-] $\mathcal{N}_1$ (Truck-only relocation 1-1): Choose random truck-only node $u$ (the node where the drone is carried by truck), and relocate it after a node $v$ in the truck tour.
	\item[-] $\mathcal{N}_2, \mathcal{N}_3$ (Truck-only relocation 2-1): Choose two random consecutive truck-only nodes $u_1, u_2$, and relocate them after a node $v$ in the truck tour as $u_1, u_2$ or $u_2, u_1$.
	\item[-] $\mathcal{N}_4$ (Truck swap 1-1):	Choose a random node $u$ in the truck tour, and swap with another node $v$ in the truck tour.
	\item[-] $\mathcal{N}_5$ (Truck swap 2-1):	Choose two random consecutive nodes $u_1, u_2$ in the truck tour such that $u_2$ does not have a drone launch or retrieval activity, and swap with another node $v$ in the truck tour. Again, we need to update the corresponding drone deliveries.
	\item[-] $\mathcal{N}_6$ (Truck swap 2-2):	Select two random consecutive nodes $u_1, u_2$ in the truck tour, and swap with two other nodes $v_1, v_2$ in the truck tour. Drone deliveries associated with those nodes are updated.
	\item[-] $\mathcal{N}_7, \mathcal{N}_8$ (Truck 2-opt): Select two random pairs of consecutive nodes $(u, x)$ and $(v, y)$ in the truck tour, and relocate them as $(u, v), (x, y)$ or $(u, y), (x, v)$.
	\item[-] $\mathcal{N}_9$ (Interdrone delivery drone-truck swap 1-1): Select a random drone node $d$, and swap it with another node $u$ in the truck tour such that $u$ is neither $d$'s launch node, rendezvous node, or the node between its launch and rendezvous.
	\item[-] $\mathcal{N}_{10}$ (Intradrone delivery drone launch swap 1-1): Select a random drone 3-tuple $\langle i, j, k \rangle$ in the drone delivery list, and swap $i$ and $j$.
	\item[-] $\mathcal{N}_{11}$ (Intradrone delivery drone rdv swap 1-1): This is similar to the above move operator, except that we swap $j$ and $k$.
	\item[-] $\mathcal{N}_{12}$ (Intradrone delivery launch rdv swap 1-1): Again, it is similar to the above move operator, but instead, we swap $i$ and $k$.
	\item[-] $\mathcal{N}_{13}$ (Drone insertion): Select a random node $j$ such that $j$ is either a truck-only node or the node in between a drone delivery, choose two other nodes $i$ and $k$ in the truck tour -- $i$ is before $k$ -- and create a new drone delivery $\langle i, j, k \rangle$. This move is only valid when no drone delivery interference exists between $i$ and $k$ or when we can say that there is no drone launch or retrieval between $i$ and $k$.
	\item[-] $\mathcal{N}_{14}$ (Drone remove): We select a random drone node $j$, remove the associated drone delivery, and reinsert $j$ between two consecutive nodes $i$ and $k$ in the truck tour.
	\item[-] $\mathcal{N}_{15}$ (Drone swap 1-1): Select two random drone deliveries $\langle i_1, j_1, k_1 \rangle$ and $\langle i_2, j_2, k_2 \rangle$, and swap $j_1$ and $j_2$. We will therefore have two new drone deliveries: $\langle i_1, j_2, k_1 \rangle$ and $\langle i_2, j_1, k_2 \rangle$.
	\item[-] $\mathcal{N}_{16}$ (Drone relocation 1-1): Select a random drone delivery $\langle i, j, k \rangle$, and choose a new launch $i'$ and rendezvous node $k'$ for $j$ to have a new drone delivery $\langle i', j, k' \rangle$.
\end{itemize}

\subsection{Restore method}
\label{section:memetic-restore}

\begin{figure}[H]
\center
\includegraphics[scale=0.38]{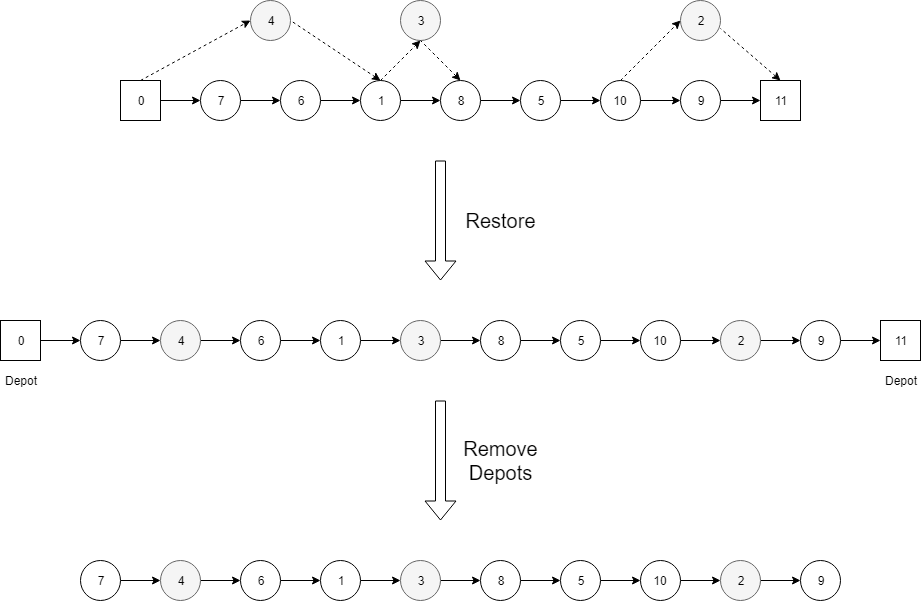}
\caption{Reinsertion in restore method. Truck travels the solid lines and drone travels the dashed lines.}\label{figure:restore-demo}
\end{figure}

To more efficiently guide the search for good solutions, a restoration method is developed in which we use the \textit{educated} TSP-D solution to update the existing giant tour individual. In detail, the new giant tour is constructed by reinserting drone nodes in the drone deliveries of the educated TSP-D solution to \textit{a random position} between their launch node and rendezvous node on the truck delivery tour of that solution. After the insertion operation is finished, two depots are removed to obtain a valid giant tour individual. As a result, we have a new giant tour individual that is formed by an ``educated'' truck tour with drone nodes being reinserted. An illustration of this process is shown in Figure \ref{figure:restore-demo}.

\subsection{Population management}
\label{section:population-management}

As an adaptation of the framework in \cite{vidal2012hybrid}, the population management mechanism in HGA remains untouched. In detail, two subpopulations are created and managed separately. They are the feasible and infeasible subpopulations. Each contains between $\mu$ to $\mu + \lambda$ individuals.

In the initialization step, $4\mu$ of individuals are created by generating a set of TSP tours using a k-cheapest insertion heuristic with k = 3 \cite{ha2018min}. The choice of a heuristic-based population comes from the analysis of \cite{murray2015flying} and the tested result in \cite{ha2018min}, which suggests the use of high-quality TSP tours instead of completely random ones. We obtain the giant tour chromosomes after these generation steps. These tours then pass through the \textit{split} method to obtain the corresponding TSP-D solutions of each individual. In the next step, individuals' TSP-D solutions are processed using the \textit{education} process to improve their qualities, and when an infeasibility occurs, they are probablistically repaired. After that, the \textit{restore method} is called to update individuals' giant tour chromosomes. The individuals are then added to the appropriate subpopulations based on their feasibilities.

Any subpopulation that exceeds the size of $\mu + \lambda$ is passed through a \textit{select survivors} method in which $\lambda$ individuals are discarded. The discarded ones are ones defined as ``clones'' or the worst individuals with respect to their biased fitnesses. Solutions are defined as clones iff they have the same giant tour (possibly in reversed order). 

Furthermore, the penalty coefficient $\omega$ is dynamically adjusted during the search for each 100 iterations. This mechanism is necessary to guide the algorithm in two search spaces. More specifically, the penalty coefficient is increased when the search produces too many infeasible solutions (meaning that it falls too deeply into the infeasible search space) and is decreased in the opposite case. In detail, let $\mathcal{E}^{REF}$ be the targeted proportion of the feasible solution, and we then adjust the parameter $\omega$ as follows: if the naturally feasible proportion is below $\mathcal{E}^{REF}$ - 5\% (is higher than $\mathcal{E}^{REF}$ + 5\%), then the penalty coefficient is increased by 1.2 (decreased by 0.85). This means that when the feasible proportion is in the range $\mathcal{E}^{REF} \pm $ 5\%, the coefficient remains unchanged to avoid the search jumping too quickly between regions in the search space.

When the search is not improved after $0.3 Iter_{NI}$ iterations, the \textit{diversification} method is called, in which we retain the best $1/3 \mu$ individuals with respect to their biased fitness and generate $4\mu$ new individuals as in the initialization phase. This technique is important because it creates new genetic materials for the search when the population has lost its diversification characteristic.

\section{Computational Results}
\label{section:computational-results}

This section presents the computational results of the HGA, which has been implemented in C++ and compiled with the ``-O3'' flag. The experiments are run on a desktop computer with an Intel Core i7-6700, 3.4 GHz processor. 

Because the parameters proposed in \cite{vidal2012hybrid} have been proven to work well on many variants of VRP, we retained most of them. In detail, the default parameters of HGA are $\mu = 15, \lambda = 25, nbElite = 6, \mathcal{E}^{REF} = 0.3, n_{close} = 0.2, \omega = 1.0, Iter_{NI} = 2500$, and $Iter_{DIV} = 0.3 \times Iter_{NI}$.

For the TSP-D parameters, we used the parameters proposed in \cite{murray2015flying}: the truck speed and drone speed were set to 40 km/h, and the drone endurance $\epsilon$ was 20 minutes. The time required to launch and retrieve the drone ($s_L$ and $s_R$) were both set to 1 minute.

As described in Section \ref{section:memetic-search-space}, there are two types of infeasibilities in a TSP-D: truck travel time and drone travel time constraint violations. From those constraint violations, we define three levels of relaxations.
\begin{itemize}
	\item \textbf{RelaxAll}: We accept both types of infeasibilities.
	\item \textbf{RelaxTruck}: We only accept the truck travel time constraint violation in infeasible solutions.
	\item \textbf{RelaxDrone}: We only accept the drone travel time constraint violation in infeasible solutions.
\end{itemize}
The impacts of these different types of relaxations are investigated in Section \ref{section:sensivity-analysis}. By default, RelaxAll is used. The default selection for the crossover is DX, which is the best performing crossover as tested in Section \ref{section:experiment-different-crossovers}.

The following sections are organized as follows. We first evaluate the performance of HGA with different instance sets and compare with the existing methods. Next, an analysis of the impacts of different crossovers is presented. Finally, we evaluate the sensitivity of each component in HGA.

\subsection{Performance on different instance sets}

In this section, we test HGA on two sets of instances: (1) 72 min-time instances of 10 customers from \cite{murray2015flying} and (2) 60 instances of 50 and 100 customers from \cite{ha2018min} under both min-time and min-cost objective functions. For the HGA, we collected its best found solutions and computed the objective function's value of solutions on average over 10 runs. Current best methods - GRASP in \cite{ha2018min} and different approaches proposed in \cite{murray2015flying} - were selected to compare with HGA. As mentioned before, the standard version of HGA with DX and RelaxAll was used in this experiment. The results for Sets (1) and (2) are presented in Tables \ref{table:compare-murray}, \ref{table:compare-grasp-cost} and \ref{table:compare-grasp-time}.

In Table \ref{table:compare-murray}, we compare HGA with the best results found by \cite{murray2015flying} and GRASP \cite{ha2018min} among 36 instances of Set (1) with two settings of drone endurance (20 and 40 minutes). The $\epsilon$ column shows the drone endurance in minutes. Column $HGA$ represents the best found solutions while column $\overline{HGA}$ reports the average values among 10 runs of our new algorithm. The values in bold text imply the best result found among the three approaches. Overall, HGA was able to improve the existing best found solutions in 9 tests and obtained results as good as the best ones in 60 tests. Column $\overline{HGA}$ shows the stability of HGA in this context when the solutions over 10 runs generally reach the best ones in all instances but two. The results also demonstrate a dominance of our HGA over GRASP in terms of solutions' quality. However, HGA is in general slower than GRASP.

Tables \ref{table:compare-grasp-cost} and \ref{table:compare-grasp-time} report the comparisons of objective value and average run time (in minutes) between HGA and GRASP in \cite{ha2018min} on instance Set (2). We collected the average value (Column ``$\overline{HGA}$'') and best solution of HGA found among repeated runs (Column ``$HGA$'') and its average run time in minutes (Column ``$T_{HGA}$''). The corresponding values of GRASP are reported in Column ``$\overline{GRASP}$'', ``$GRASP$'', and ``$T_{GRASP}$''. Column ``Change(\%)'' calculates the percentage change between best found objective values of HGA and GRASP. A negative value indicates an improvement of our new method in comparison to GRASP. With respect to this comparison, HGA shows improvements in terms of solutions` quality in both min-cost and min-time objectives. 

In detail, for min-cost TSP-D (Table \ref{table:compare-grasp-cost}), the average objective values of solutions of HGA are even better than those of the best found solutions of GRASP on most instances (see Columns ``$\overline{HGA}$''and ``$GRASP$'').  The proposed algorithm can significantly improve existing best known solutions by 6.16\% and 15.10\% on average (up to nearly 15\% and 20\%) for 50- and 100-customer instances, respectively. We can observe that the algorithm performs better in large instances (i.e., 100-customer instances). However, it is worth mentioning that GRASP performs better on two instances D5 and D6.  Regarding run time, HGA is 1.5 to 2 times slower than GRASP due to its more complex design. This result is acceptable since it still can deliver significantly better results in less than 1 minute for 50-customer instances and less than 5 minutes for 100-customer instances. 

For min-time TSP-D (Table \ref{table:compare-grasp-time}), HGA can also improve the existing best known solutions found by GRASP on all instances but not as significantly as in min-cost TSP-D. In detail, the improvements are 2.39\% and 4.05\% on average (and up to nearly 6\% and 8\%) for 50- and 100-customer instances, respectively. Again, HGA performs approximately 1.5 times slower than GRASP but can still deliver better solutions in less than 1 minute and 5 minutes for 50- and 100-customer instances, respectively.

\begin{table}[H]
\centering
\scalebox{0.6} {
\begin{tabular}{|c|c|c|c|c|c||c|c|c|c|c|c|}
\hline
Instance & $\epsilon$  & Murray et al. & $GRASP$  & $HGA$ & $\overline{HGA}$  & Instance & $\epsilon$  & Murray et al & $GRASP$  & $HGA$ & $\overline{HGA}$ \\ \hline
437v1  & 20 & 56.468          & 57.446 & 56.468          & 56.468 & 440v7  & 20 & 49.996          & 49.776 & \textbf{49.422} & 49.422 \\
437v1  & 40 & 50.573          & 50.573 & 50.573          & 50.573 & 440v7  & 40 & 49.204          & 49.204 & 49.204          & 49.204 \\
437v2  & 20 & 53.207          & 53.207 & 53.207          & 53.207 & 440v8  & 20 & 62.796          & 62.700 & 62.576          & 62.576 \\
437v2  & 40 & 47.311          & 47.311 & 47.311          & 47.311 & 440v8  & 40 & 62.270          & 62.004 & 62.004          & 62.004 \\
437v3  & 20 & 53.687          & 54.664 & 53.687          & 53.687 & 440v9  & 20 & 42.799          & 42.566 & \textbf{42.533} & 42.533 \\
437v3  & 40 & 53.687          & 53.687 & 53.687          & 53.687 & 440v9  & 40 & 42.799          & 42.566 & \textbf{42.533} & 42.533 \\
437v4  & 20 & 67.464          & 67.464 & 67.464          & 67.464 & 440v10 & 20 & 43.076          & 43.076 & 43.076          & 43.076 \\
437v4  & 40 & 66.487          & 66.487 & 66.487          & 66.487 & 440v10 & 40 & 43.076          & 43.076 & 43.076          & 43.076 \\
437v5  & 20 & 50.551          & 50.551 & 50.551          & 50.551 & 440v11 & 20 & 49.204          & 49.204 & 49.204          & 49.204 \\
437v5  & 40 & 45.835          & 44.835 & 44.835          & 44.835 & 440v11 & 40 & 49.204          & 49.204 & 49.204          & 49.204 \\
437v6  & 20 & \textbf{45.176} & 47.601 & 47.311          & 47.311 & 440v12 & 20 & 62.004          & 62.004 & 62.004          & 62.004 \\
437v6  & 40 & 45.863          & 43.602 & 43.602          & 43.602 & 440v12 & 40 & 62.004          & 62.004 & 62.004          & 62.004 \\
437v7  & 20 & 49.581          & 49.581 & 49.581          & 49.581 & 443v1  & 20 & 69.586          & 69.586 & 69.586          & 69.586 \\
437v7  & 40 & 46.621          & 46.621 & 46.621          & 46.621 & 443v1  & 40 & 55.493          & 55.493 & 55.493          & 55.493 \\
437v8  & 20 & 62.381          & 62.381 & 62.381          & 62.381 & 443v2  & 20 & 72.146          & 72.146 & 72.146          & 72.146 \\
437v8  & 40 & 59.776          & 59.416 & 59.416          & 59.416 & 443v2  & 40 & 58.053          & 58.053 & 58.053          & 58.053 \\
437v9  & 20 & 45.985          & 42.945 & \textbf{42.416} & 42.416 & 443v3  & 20 & 77.344          & 77.344 & 77.344          & 77.344 \\
437v9  & 40 & 42.416          & 42.416 & 42.416          & 42.416 & 443v3  & 40 & 69.175          & 68.431 & 68.431          & 68.431 \\
437v10 & 20 & 42.416          & 41.729 & 41.729          & 41.729 & 443v4  & 20 & 90.144          & 90.144 & 90.144          & 90.144 \\
437v10 & 40 & 41.729          & 41.729 & 41.729          & 41.729 & 443v4  & 40 & 82.700          & 83.700 & 82.700          & 82.700 \\
437v11 & 20 & 42.896          & 42.896 & 42.896          & 42.896 & 443v5  & 20 & 55.493          & 58.210 & \textbf{54.973} & 55.077 \\
437v11 & 40 & 42.896          & 42.896 & 42.896          & 42.896 & 443v5  & 40 & 53.447          & 51.929 & 51.929          & 51.929 \\
437v12 & 20 & 56.696          & 56.425 & \textbf{56.273} & 56.273 & 443v6  & 20 & 58.053          & 58.053 & \textbf{55.209} & 55.209 \\
437v12 & 40 & 55.696          & 55.696 & 55.696          & 55.696 & 443v6  & 40 & 52.329          & 52.329 & 52.329          & 52.329 \\
440v1  & 20 & 49.430          & 50.164 & 49.430          & 49.430 & 443v7  & 20 & \textbf{64.409} & 65.523 & 65.523          & 65.523 \\
440v1  & 40 & 46.886          & 46.886 & 46.886          & 46.886 & 443v7  & 40 & 60.743          & 60.743 & 60.743          & 60.743 \\
440v2  & 20 & 50.708          & 51.828 & 50.708          & 50.708 & 443v8  & 20 & \textbf{77.209} & 78.323 & 78.323          & 78.323 \\
440v2  & 40 & 46.423          & 46.423 & 46.423          & 46.423 & 443v8  & 40 & 73.967          & 72.967 & 72.967          & 72.967 \\
440v3  & 20 & 56.102          & 58.502 & 56.102          & 56.102 & 443v9  & 20 & 49.049          & 45.931 & 45.931          & 45.931 \\
440v3  & 40 & 53.933          & 53.933 & 53.933          & 53.933 & 443v9  & 40 & 47.250          & 45.931 & 45.931          & 45.931 \\
440v4  & 20 & 69.902          & 73.091 & 69.902          & 69.902 & 443v10 & 20 & 47.935          & 46.935 & 46.935          & 46.935 \\
440v4  & 40 & 68.397          & 68.397 & 68.397          & 68.397 & 443v10 & 40 & 47.935          & 46.935 & 46.935          & 46.935 \\
440v5  & 20 & 43.533          & 44.624 & 43.533          & 43.533 & 443v11 & 20 & 57.382          & 56.395 & 56.395          & 56.395 \\
440v5  & 40 & 43.533          & 43.533 & 43.533          & 43.533 & 443v11 & 40 & 56.395          & 56.395 & 56.395          & 56.395 \\
440v6  & 20 & 44.076          & 44.122 & \textbf{43.949} & 43.949 & 443v12 & 20 & 69.195          & 69.195 & 69.195          & 69.195 \\
440v6  & 40 & 44.076          & 43.944 & \textbf{43.810} & 43.853 & 443v12 & 40 & 69.195          & 69.195 & 69.195          & 69.195 \\ \hline
\end{tabular}
}
\caption{Comparison set 1 instances under Min-time objective}
\label{table:compare-murray}
\end{table}

\begin{sidewaystable}[]
\centering
\scalebox{0.7} {
\begin{tabular}{|c|c|c|c|c|c|c|c||c|c|c|c|c|c|c|c|}
\hline
Instance & $GRASP$ & $\overline{GRASP}$ & $T_{GRASP}$ (min) & $HGA$ & $\overline{HGA}$ & Change(\%)        & $T_{HGA}$ (min) & Instance & $GRASP$ & $\overline{GRASP}$ & $T_{GRASP}$ (min) & HGA & $\overline{HGA}$ & Change(\%)      & $T_{HGA}$ (min) \\ \hline
B1  & 1372.82 & 1413.24 & 0.27 & 1225.78 & 1239.85 & \textbf{-10.71} & 0.5  & E1  & 2206.53 & 2255.99 & 2.28 & 1775.1  & 1802.47 & \textbf{-19.55} & 3.47 \\
B2  & 1491.3  & 1513.98 & 0.26 & 1381.89 & 1402.98 & \textbf{-7.34}  & 0.38 & E2  & 2210.61 & 2273.09 & 2.28 & 1795.03 & 1830.95 & \textbf{-18.80} & 3.66 \\
B3  & 1503.78 & 1521.67 & 0.28 & 1357.17 & 1370.82 & \textbf{-9.75}  & 0.45 & E3  & 2248.16 & 2312.76 & 2.48 & 1818.16 & 1861.59 & \textbf{-19.13} & 3.5  \\
B4  & 1396.17 & 1426.2  & 0.27 & 1282.16 & 1292.87 & \textbf{-8.17}  & 0.49 & E4  & 2179.06 & 2223.97 & 2.97 & 1776.58 & 1822.36 & \textbf{-18.47} & 3.68 \\
B5  & 1457.91 & 1500.9  & 0.31 & 1351.37 & 1357.61 & \textbf{-7.31}  & 0.4  & E5  & 2286.16 & 2360.3  & 2.87 & 1866.22 & 1899.25 & \textbf{-18.37} & 3.55 \\
B6  & 1316.08 & 1353.76 & 0.27 & 1159.79 & 1174.30 & \textbf{-11.88} & 0.44 & E6  & 2244.62 & 2313.86 & 3.26 & 1795.17 & 1831.23 & \textbf{-20.02} & 4.12 \\
B7  & 1370.05 & 1399.71 & 0.24 & 1308.25 & 1322.70 & \textbf{-4.51}  & 0.42 & E7  & 2249.09 & 2313.67 & 3.18 & 1892.41 & 1923.18 & \textbf{-15.86} & 3.46 \\
B8  & 1484.93 & 1517.23 & 0.25 & 1255.61 & 1275.61 & \textbf{-15.44} & 0.62 & E8  & 2220.88 & 2272.55 & 3.15 & 1813.73 & 1831.46 & \textbf{-18.33} & 4.3  \\
B9  & 1442.09 & 1468.86 & 0.28 & 1355.32 & 1363.53 & \textbf{-6.02}  & 0.52 & E9  & 2279.91 & 2326.29 & 2.87 & 1882.71 & 1901.76 & \textbf{-17.42} & 5.01 \\
B10 & 1392.54 & 1429.57 & 0.25 & 1252.9  & 1257.48 & \textbf{-10.03} & 0.47 & E10 & 2324.74 & 2384.52 & 3.41 & 1870.55 & 1932.75 & \textbf{-19.54} & 4.07 \\
C1  & 2870.41 & 2935.87 & 0.21 & 2679.1  & 2703.14 & \textbf{-6.66}  & 0.29 & F1  & 4569.83 & 4648.2  & 1.85 & 3766.63 & 3854.56 & \textbf{-17.58} & 3.24 \\
C2  & 2804.47 & 2868.67 & 0.26 & 2750.74 & 2755.04 & \textbf{-1.92}  & 0.36 & F2  & 4186.76 & 4318.78 & 2.38 & 3469.54 & 3575.90 & \textbf{-17.13} & 3.95 \\
C3  & 3087.55 & 3185.09 & 0.16 & 2932.78 & 2952.48 & \textbf{-5.01}  & 0.32 & F3  & 4414.38 & 4563.64 & 2.45 & 3751.07 & 3891.27 & \textbf{-15.03} & 3.02 \\
C4  & 2844.1  & 2916.86 & 0.20 & 2655.25 & 2676.82 & \textbf{-6.64}  & 0.67 & F4  & 4499.09 & 4600.27 & 2.14 & 3818.62 & 3862.57 & \textbf{-15.12} & 3.27 \\
C5  & 3323.92 & 3367.34 & 0.19 & 3133.69 & 3156.88 & \textbf{-5.72}  & 0.36 & F5  & 4381.37 & 4597.32 & 2.66 & 3756.78 & 3807.86 & \textbf{-14.26} & 3.28 \\
C6  & 3433.99 & 3472.39 & 0.19 & 3238.92 & 3268.49 & \textbf{-5.68}  & 0.4  & F6  & 4032.9  & 4171.8  & 2.63 & 3465.56 & 3560.66 & \textbf{-14.07} & 2.54 \\
C7  & 3001.13 & 3047.71 & 0.21 & 2681.01 & 2738.71 & \textbf{-10.67} & 0.91 & F7  & 4076.31 & 4213.52 & 2.84 & 3601.78 & 3660.90 & \textbf{-11.64} & 4.83 \\
C8  & 3481.17 & 3557.99 & 0.22 & 3250.19 & 3259.22 & \textbf{-6.64}  & 0.91 & F8  & 4491.2  & 4597.9  & 2.77 & 3803.14 & 3930.37 & \textbf{-15.32} & 3.91 \\
C9  & 3267.23 & 3306.38 & 0.19 & 3032.73 & 3056.08 & \textbf{-7.18}  & 0.58 & F9  & 4388.91 & 4463.39 & 2.55 & 3873.51 & 3904.18 & \textbf{-11.74} & 4.07 \\
C10 & 3291.2  & 3356.29 & 0.23 & 3082.07 & 3117.05 & \textbf{-6.35}  & 0.61 & F10 & 4173.64 & 4567.84 & 2.57 & 3837.47 & 3895.06 & \textbf{-8.05}  & 3.1  \\
D1  & 4159.39 & 4389.24 & 0.21 & 3927.97 & 3928.12 & \textbf{-5.56}  & 0.43 & G1  & 5947.97 & 6148.5  & 1.94 & 5084.56 & 5312.36 & \textbf{-14.52} & 3.08 \\
D2  & 4275.46 & 4334.4  & 0.19 & 4097    & 4113.47 & \textbf{-4.17}  & 0.58 & G2  & 5882.97 & 5987.64 & 2.63 & 5198.89 & 5234.67 & \textbf{-11.63} & 3.16 \\
D3  & 4085.71 & 4191.08 & 0.18 & 3846.84 & 3861.61 & \textbf{-5.85}  & 0.39 & G3  & 6074.57 & 6138.94 & 2.82 & 5063.87 & 5116.38 & \textbf{-16.64} & 4.08 \\
D4  & 4612.46 & 4714.62 & 0.21 & 4334.32 & 4334.32 & \textbf{-6.03}  & 0.31 & G4  & 6458.96 & 6632.14 & 2.39 & 5542.19 & 5703.23 & \textbf{-14.19} & 4.1  \\
D5  & 4717.67 & 4793.39 & 0.20 & 4732.96 & 4745.25 & \textbf{0.32}   & 0.65 & G5  & 6198.95 & 6329.25 & 2.59 & 5496.77 & 5566.09 & \textbf{-11.33} & 2.94 \\
D6  & 4405.02 & 4485.87 & 0.20 & 4546.99 & 4604.40 & \textbf{3.22}   & 0.77 & G6  & 6049.34 & 6343.26 & 2.95 & 5377.51 & 5463.62 & \textbf{-11.11} & 2.67 \\
D7  & 4749.57 & 4796.23 & 0.25 & 4634.53 & 4657.77 & \textbf{-2.42}  & 0.73 & G7  & 5889.08 & 6023.11 & 2.85 & 5318.17 & 5396.12 & \textbf{-9.69}  & 4.98 \\
D8  & 4143.03 & 4287.87 & 0.20 & 3911.94 & 3963.02 & \textbf{-5.58}  & 0.64 & G8  & 5599.55 & 5871.96 & 2.62 & 5112.58 & 5246.50 & \textbf{-8.70}  & 5.03 \\
D9  & 4653.73 & 4688.16 & 0.22 & 4469.78 & 4490.67 & \textbf{-3.95}  & 0.66 & G9  & 6050.8  & 6254.5  & 3.08 & 4996.42 & 5187.45 & \textbf{-17.43} & 4.3  \\
D10 & 4260.6  & 4301.83 & 0.20 & 4208.93 & 4232.38 & \textbf{-1.21}  & 0.95 & G10 & 6249.69 & 6534.13 & 2.70 & 5473.91 & 5598.25 & \textbf{-12.41} & 3.77  \\ \hline
Mean & & & 0.22 & & & \textbf{-6.16} & 0.51 & & & & 2.65 & & & \textbf{-15.10} & 3.68 \\ \hline
\end{tabular}
}
\caption{Comparison with GRASP under Min-cost objective - Instance set 2}
\label{table:compare-grasp-cost}
\end{sidewaystable}

\begin{sidewaystable}[]
\centering
\scalebox{0.7} {
\begin{tabular}{|c|c|c|c|c|c|c|c||c|c|c|c|c|c|c|c|}
\hline
Instance & $GRASP$ & $\overline{GRASP}$ & $T_{GRASP}$ (min) & $HGA$ & $\overline{HGA}$ & Change(\%) & $T_{HGA}$ (min) & Instance & $GRASP$ & $\overline{GRASP}$ & $T_{GRASP}$ (min) & $HGA$ & $\overline{HGA}$ & Change(\%)      & $T_{HGA}$ (min) \\ \hline
B1  & 120.68 & 121.69 & 0.45 & 115.65 & 116.43 & \textbf{-4.17} & 0.76 & E1  & 188.58 & 192.08 & 5.45 & 187.67 & 188.32 & \textbf{-0.48} & 3.6  \\
B2  & 118.53 & 119.46 & 0.48 & 118.39 & 118.39 & \textbf{-0.12} & 0.33 & E2  & 190.55 & 192.88 & 5.71 & 187.21 & 188.01 & \textbf{-1.75} & 5.6  \\
B3  & 119.7  & 120.25 & 0.52 & 116.21 & 116.39 & \textbf{-2.92} & 0.57 & E3  & 189.05 & 192.83 & 5.65 & 188.09 & 188.89 & \textbf{-0.51} & 4.58 \\
B4  & 123.02 & 124.7  & 0.36 & 118.71 & 119.26 & \textbf{-3.50} & 0.47 & E4  & 188.61 & 191.27 & 4.54 & 186.23 & 186.99 & \textbf{-1.26} & 4.69 \\
B5  & 119.46 & 120.77 & 0.48 & 115.78 & 115.91 & \textbf{-3.08} & 0.58 & E5  & 190.47 & 193.61 & 4.34 & 187.71 & 188.26 & \textbf{-1.45} & 4.06 \\
B6  & 119.54 & 121.46 & 0.39 & 114.31 & 115.46 & \textbf{-4.38} & 0.88 & E6  & 190.32 & 193.86 & 4.10 & 189.16 & 189.44 & \textbf{-0.61} & 4.84 \\
B7  & 118.54 & 121.02 & 0.33 & 115.52 & 115.63 & \textbf{-2.55} & 0.62 & E7  & 191.51 & 194.41 & 4.33 & 190.39 & 190.89 & \textbf{-0.58} & 3.84 \\
B8  & 119.36 & 119.99 & 0.35 & 117.9  & 118.04 & \textbf{-1.22} & 0.78 & E8  & 190.47 & 193.74 & 3.86 & 189.02 & 189.54 & \textbf{-0.76} & 4.22 \\
B9  & 118.26 & 119.86 & 0.42 & 117.64 & 117.72 & \textbf{-0.52} & 0.39 & E9  & 191.12 & 193.7  & 4.31 & 189.76 & 189.94 & \textbf{-0.71} & 4    \\
B10 & 119.8  & 121.27 & 0.37 & 117.38 & 117.70 & \textbf{-2.02} & 0.6  & E10 & 189.71 & 193.28 & 4.17 & 189.45 & 189.91 & \textbf{-0.14} & 3.4  \\
C1  & 220.63 & 222.6  & 0.27 & 215.07 & 215.37 & \textbf{-2.52} & 0.6  & F1  & 341.68 & 344.98 & 2.65 & 322.94 & 326.10 & \textbf{-5.48} & 5.73 \\
C2  & 210.39 & 211.14 & 0.41 & 209.23 & 210.11 & \textbf{-0.55} & 0.53 & F2  & 325.7  & 330.63 & 2.87 & 308.74 & 310.89 & \textbf{-5.21} & 5.24 \\
C3  & 214.61 & 215.31 & 0.28 & 212.02 & 212.22 & \textbf{-1.21} & 0.38 & F3  & 336.35 & 340.92 & 3.88 & 309.67 & 313.55 & \textbf{-7.93} & 5.61 \\
C4  & 225.15 & 225.47 & 0.25 & 212.08 & 213.27 & \textbf{-5.81} & 0.6  & F4  & 326.79 & 334.69 & 1.98 & 311.37 & 314.96 & \textbf{-4.72} & 6.06 \\
C5  & 226.36 & 233.97 & 0.32 & 223.06 & 224.57 & \textbf{-1.46} & 0.48 & F5  & 335.88 & 344.61 & 2.04 & 314.82 & 317.83 & \textbf{-6.27} & 6.57 \\
C6  & 240.37 & 242.22 & 0.27 & 234.01 & 235.56 & \textbf{-2.65} & 0.31 & F6  & 309.71 & 319.22 & 2.23 & 294.38 & 297.47 & \textbf{-4.95} & 4.7  \\
C7  & 227.73 & 229.56 & 0.20 & 222.27 & 223.40 & \textbf{-2.40} & 0.51 & F7  & 317.86 & 330.83 & 1.67 & 311.41 & 316.15 & \textbf{-2.03} & 4.92 \\
C8  & 242.19 & 245.12 & 0.37 & 234.26 & 237.53 & \textbf{-3.27} & 0.46 & F8  & 345.44 & 350.75 & 1.96 & 323.74 & 326.40 & \textbf{-6.28} & 5.21 \\
C9  & 237.98 & 241.07 & 0.26 & 226.01 & 227.43 & \textbf{-5.03} & 0.68 & F9  & 339.53 & 342.41 & 1.83 & 315.56 & 318.47 & \textbf{-7.06} & 4.66 \\
C10 & 230.03 & 235.47 & 0.33 & 226.17 & 226.17 & \textbf{-1.68} & 0.48 & F10 & 332.05 & 340.35 & 1.87 & 312.7  & 315.13 & \textbf{-5.83} & 3.94 \\
D1  & 315.8  & 318.9  & 0.31 & 306.39 & 307.09 & \textbf{-2.98} & 0.61 & G1  & 437.48 & 450.23 & 1.87 & 417.92 & 425.19 & \textbf{-4.47} & 4.45 \\
D2  & 317.15 & 322.85 & 0.29 & 313.93 & 315.64 & \textbf{-1.02} & 0.57 & G2  & 415.32 & 424.88 & 2.67 & 389.64 & 390.14 & \textbf{-6.18} & 2.4  \\
D3  & 300.4  & 303.26 & 0.29 & 295.86 & 297.54 & \textbf{-1.51} & 0.6  & G3  & 446.6  & 454.98 & 2.45 & 411.47 & 415.14 & \textbf{-7.87} & 4.9  \\
D4  & 333.47 & 336.7  & 0.31 & 323.72 & 324.60 & \textbf{-2.92} & 0.56 & G4  & 449.68 & 465.83 & 1.83 & 433.09 & 435.56 & \textbf{-3.69} & 4.67 \\
D5  & 324.68 & 326.46 & 0.25 & 321.46 & 321.83 & \textbf{-0.99} & 0.4  & G5  & 434.6  & 446.73 & 1.67 & 421.05 & 422.49 & \textbf{-3.12} & 4.48 \\
D6  & 315.16 & 317.73 & 0.28 & 313.21 & 313.65 & \textbf{-0.62} & 0.49 & G6  & 450.28 & 462    & 1.73 & 415.46 & 420.84 & \textbf{-7.73} & 5.51 \\
D7  & 329.31 & 330.24 & 0.33 & 316.65 & 317.83 & \textbf{-3.84} & 0.32 & G7  & 420    & 439.62 & 1.42 & 409.31 & 412.14 & \textbf{-2.55} & 5.21 \\
D8  & 306.28 & 312.12 & 0.31 & 293.76 & 296.51 & \textbf{-4.09} & 0.58 & G8  & 442.67 & 453.19 & 1.71 & 406.51 & 407.89 & \textbf{-8.17} & 5.08 \\
D9  & 326.09 & 331.31 & 0.27 & 317.85 & 318.31 & \textbf{-2.53} & 0.41 & G9  & 456.78 & 469.49 & 1.32 & 428.16 & 435.75 & \textbf{-6.27} & 5.91 \\
D10 & 306.1  & 309.54 & 0.29 & 305.51 & 305.54 & \textbf{-0.19} & 0.41 & G10 & 460.89 & 470.44 & 2.15 & 426.82 & 430.94 & \textbf{-7.39} & 5.4    \\ \hline
Mean & & & 0.33 & & & \textbf{-2.39} & 0.52 & & & & 2.66 & & & \textbf{-4.05} & 4.51 \\ \hline
\end{tabular}
}
\caption{Comparison with GRASP under Min-time objective - Instance set 2}
\label{table:compare-grasp-time}
\end{sidewaystable}

\subsection{Performance under different crossovers}
\label{section:experiment-different-crossovers}

\begin{figure}[H]
\center
\includegraphics[scale=0.45]{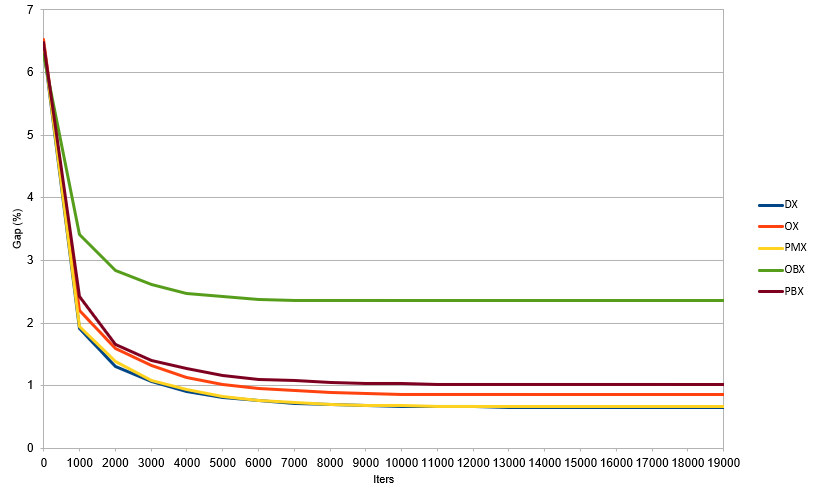}
\caption{Crossovers' performance - Min-Time objective.}\label{figure:converge-time}
\end{figure}

\begin{figure}[H]
\center
\includegraphics[scale=0.45]{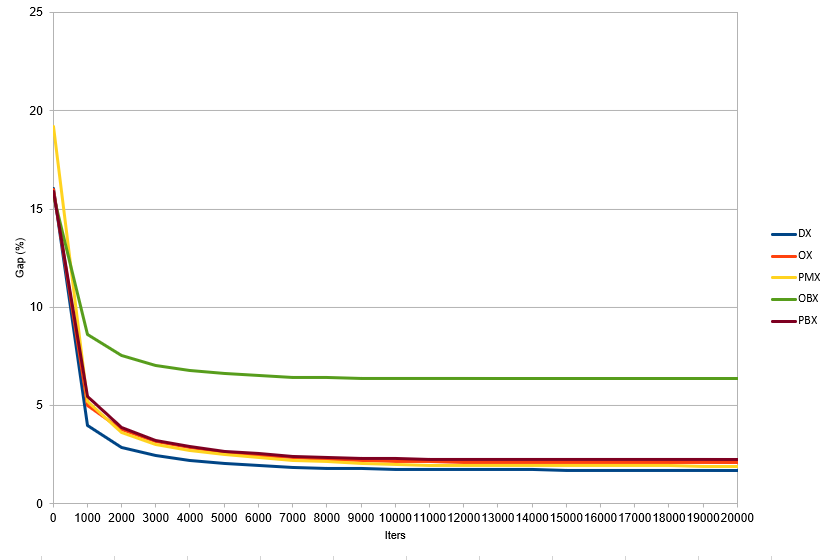}
\caption{Crossovers' performance - Min-Cost objective.}\label{figure:converge-cost}
\end{figure}

We evaluate the performance of HGA when using our proposed crossover over 4 classical crossovers \cite{potvin1996genetic} (OX, PMX, OBX, and PBX) in Tables \ref{table:crossovers} under two objectives with instance Set (2) mentioned in the above section. Again, HGA was repeatedly run 10 times for each choice of crossover, and we have conducted 6000 tests in total. For each crossover, we report the average percentage gap with the best found solution (regardless of crossover), the run time in minutes (Column ``$T$''), the standard deviation (Column ``sd'') and the geometric mean value (row ``Mean''). Furthermore, a comparison of the convergence of these crossovers in both objectives is presented in Figures \ref{figure:converge-time} and \ref{figure:converge-cost}, where the Y-axis shows the averaged percentage gap with the best found solutions, and the X-axis contains the maximum number of iterations over which an improvement could be made.

Overall, DX delivers the best value among other crossovers in terms of percentage gap. For min-cost, DX is approximately 18\%, 5.7\%, 283\%, and 16.5\% better than OX, PMX, OBX, and PBX, respectively. For min-time, that superiority is approximately 26.5\%, 10.2\%, 283\%, and 46.9\%. As can be seen, OBX performs worst among the crossovers, possibly due to its design, for which only a random number of separated nodes is copied from the parent. This causes the OBX to have a smaller chance of transmitting ``good'' materials from its parent such as good drone deliveries or good, complete truck deliveries. The performances of OX and PMX, on the other hand, were much closer to those of DX, especially for PMX in the min-cost problem, being only 5.7\% inferior. This result is because OX and PMX are both designed to copy a random subsequence of the parent to the children, thus having a high chance of transmitting ``good'' materials such as complete drone or truck deliveries from parent to offspring.

With respect to run time, OBX performs nearly 1.5 to 2 times faster than other crossovers. However, due to its poor performance, this fast run time is not valuable. Other crossovers deliver similar run times -- less than 2 minutes on average -- which is an acceptable value. 

When considering standard deviation, DX, OX, PMX and PBX perform stably, the values of which are mostly less than 0.5\% and no more than 1\%, while OBX shows its instability in delivering values that are more than 0.5\% and up to nearly 1.3\%.

Finally, from Figures \ref{figure:converge-time} and \ref{figure:converge-cost}, we can see a similar pattern in the convergences of all the crossovers. They all converge quickly in the first 5000 iterations. 

\begin{table}[H]
\centering
\scalebox{0.55} {
\begin{tabular}{|c|c|c|c|c|c|c|c|c|c|c|c|c|c|c|c|}
\hline
 & \multicolumn{3}{|c|}{DX} & \multicolumn{3}{|c|}{OX} & \multicolumn{3}{|c|}{PMX} & \multicolumn{3}{|c|}{OBX} & \multicolumn{3}{|c|}{PBX} \\ \hline
          & Gap   & $T$ (min)   & sd   & Gap   & $T$ (min)   & sd   & Gap   & $T$ (min)   & sd    & Gap    & $T$ (min)  & sd    & Gap   & $T$ (min)   & sd    \\ \hline
Min-Cost   & \textbf{1.39} & 1.37 & 0.86 & 1.64 & 1.44 & 0.87 & 1.47 & 1.31 & 0.92 & 5.33 & 0.87 & 1.28 & 1.62 & 1.53 & 0.95 \\ \hline
Min-Time   & \textbf{0.49} & 1.55 & 0.33 & 0.62 & 1.50 & 0.41 & 0.54 & 1.48 & 0.30 & 1.88  & 0.96 & 0.51 & 0.72 & 1.73 & 0.40 \\ \hline
\end{tabular}
}
\caption{Crossover performance comparison - Min-cost and Min-time objective}
\label{table:crossovers}
\end{table}

\subsection{Sensitivity analyses}
\label{section:sensivity-analysis}
This section provides analyses, as shown in Table \ref{table:sensivity}, of the impact of the key components of HGA under the measurement of percentage gap on average of solutions over 10 runs to the best known solutions (BKS). The investigated components are the restore method, repair mechanism, relaxation choice (relax truck/drone endurance checking one by one), infeasibility of solutions and diversity contribution. We adapted the standard setting (crossover DX is used with parameters mentioned at the beginning of Section \ref{section:computational-results}) and modified each of the key components to test their impact. In detail, we have the following.
\begin{itemize}
	\item \textbf{No INF}: Instead of relaxing the endurance constraint on truck and drone travel times, we insist that it hold. Therefore, no infeasible solution is allowed.
	\item \textbf{No DIV}: We do not count the diversity contribution (setting it to 0) during the calculation of biased fitness.
	\item \textbf{No REPAIR}: We do not use a repair method in HGA.
	\item \textbf{No RESTORE}: We do not use a restore method in HGA.
	\item \textbf{RelaxTruck}: We only allow for infeasible solutions in which the endurance constraint is violated by truck travel times but not the drone's time.
	\item \textbf{RelaxDrone}: Opposite \textbf{RelaxTruck}, where we only allow for violation of drone travel time.
\end{itemize}

The experiment results show that HGA is indeed sensible to its parameters (infeasibility, diversity contribution, repair, and restore method) in such a way that any change to those values negatively impact the algorithm's performance. However, those negative changes do not share the same impact. In detail, eliminating the role of the restore method (\textbf{No RESTORE}) strongly reduces the performance of HGA, which proves the necessity of this problem-specific component to the general framework in order to efficiently solve the TSP-D problem.

The infeasible solutions management, diversity contribution and repair mechanism (\textbf{No INF}, \textbf{No DIV} and \textbf{No REPAIR}) also contribute to the performance of HGA, notably the \textbf{No INF} and \textbf{No DIV}, where the increment compared to the standard gap exceeds 50\%. This result proves the effectiveness of using both feasible and infeasible solutions as well as the importance of a diversity control mechanism to avoid the search becoming stuck too quickly in the local minima.

Regarding the relaxation selection (\textbf{RelaxTruck}, \textbf{RelaxDrone}), we can observe the negative impact of these choices on the performance of HGA for both objectives. However, this impact is not the same for each of the objective types. In detail, while the min-cost objective performs well when the drone travel time constraint is relaxed (\textbf{RelaxDrone}), the min-cost objective delivers a gap close to the standard gap when the truck travel time constraint is relaxed (\textbf{RelaxTruck}). This phenomenon could be explained as follows.

In the min-cost problem, the longer the distance (or time) the truck travels between launch and rendezvous nodes during a drone delivery is, the greater the impact on the travel cost it would receive, as the transportation cost of the truck is many times larger than that of the drone. Hence, with the \textbf{RelaxTruck} option for which the truck travel time constraint is relaxed and the drone travel time constraint is imposed, the truck would be less likely to receive this relaxation advantage because of its high transportation cost per unit distance. On the other hand, when the drone travel time constraint is not enforced (\textbf{RelaxDrone}), the algorithm could have infeasible solutions in which the drone will take the longer arcs (because of its small transportation cost). These solutions then have more opportunities to be repaired to become a high quality solution.

In the min-time problem, as analysed in \cite{ha2018min}, the frequency at which the drone is used is much less than that in the min-cost problem. Therefore, min-time solution quality depends more on truck tour quality. Hence, when the truck travel time constraint is relaxed (\textbf{RelaxTruck}), we could have infeasible solutions in which the drone arrives at the rendezvous node before the truck. This is the ideal situation for the truck as it could immediately proceed to the next customer location or prepare a parcel for the next launch without waiting for the drone to arrive \cite{murray2015flying}. This could shorten the truck's wait time and possibly lead to a good truck tour. Thus, along with the repair method, these kinds of infeasible solutions would have more chances to be repaired to become a high quality solution. On the other hand, the opposite fact occurs when the drone travel time constraint is relaxed (\textbf{RelaxDrone}), meaning that the truck is more likely to wait for the drone at the rendezvous node, therefore having less chance of obtaining good solutions.

\begin{table}[H]
\centering
\scalebox{0.75} {
\begin{tabular}{|c|c|c|c|c|c|c|c|}
\hline
     & No INF & No DIV & No REPAIR & No RESTORE & RelaxTruck & RelaxDrone & Standard \\ \hline
Min-cost & 2.39   & 2.19   & 1.34      & 5.42       & 2.19       & 1.30       & 1.29     \\
Min-time & 0.84   & 0.94   & 0.58      & 1.39       & 0.64       & 0.79       & 0.53     \\ \hline
\end{tabular}
}
\caption{Sensivity analysis of key components}
\label{table:sensivity}
\end{table}

\section{Conclusion}
\label{section:conclusion}

In this paper, we presented a new hybrid genetic algorithm -- HGA -- to effectively solve the TSP-D under both min-cost and min-time objectives. Our algorithm includes new problem-tailored components such as local searches, crossover, restore method and penalized mechanism to effectively guide the search for good solutions. Computational experiments show that HGA outperforms the existing methods in terms of solution quality to become the stat-of-the-art approximation method proposed for the TSP-D problems. Our method can also improve a number of the best known solutions found in the literature. An extensive analysis was carried out to demonstrate the importance of the new components to the overall performance of HGA. In future work, we intend to develop an efficient exact method to better investigate the performance of the algorithm. Also, we would like to test HGA on other variants of the TSP-D such as the version with multiple trucks and multiple drones under both objectives.

\section*{Acknowledgement}
This research is funded by Vietnam National Foundation for Science and
Technology Development (NAFOSTED) under grant number Grant Number 102.99-2016.21.

\section*{References}

\bibliography{memetic}

\end{document}